\documentclass[letterpaper]{article} 
\usepackage{aaai25}   
\usepackage{times}  
\usepackage{helvet}  
\usepackage{courier}  
\usepackage[hyphens]{url}  
\usepackage{graphicx} 
\usepackage{natbib}  
\usepackage{caption} 
\usepackage{algorithm}
\usepackage{algorithmic}
\usepackage{multirow}
\usepackage{booktabs}
\usepackage{amsmath}
\usepackage{overpic} 
\usepackage{float}   
\usepackage{inconsolata}
\usepackage{microtype}
\usepackage{times}
\usepackage{bbding}
\usepackage{latexsym}
\usepackage{subcaption} 
\usepackage{graphicx}   

\usepackage{newfloat}
\usepackage{listings}
\DeclareCaptionStyle{ruled}{labelfont=normalfont,labelsep=colon,strut=off} 
\floatstyle{ruled}
\newfloat{listing}{tb}{lst}{}
\floatname{listing}{Listing}
\title{SEE: Sememe Entanglement Encoding for Transformer-based Models Compression}

\author{
    Jing Zhang\textsuperscript{\rm 1},
    Shuzhen Sun\textsuperscript{\rm 1},
    Peng Zhang\textsuperscript{\rm 1}\thanks{Corresponding author: Peng Zhang (pzhang@tju.edu.cn)},
    Guangxing Cao\textsuperscript{\rm 1},
    Hui Gao\textsuperscript{\rm 1},
    Xindian Ma\textsuperscript{\rm 1},
    Nan Xu\textsuperscript{\rm 2},
    Yuexian Hou\textsuperscript{\rm 1}\thanks{Corresponding author: Yuexian Hou (yxhou@tju.edu.cn)}
}
\affiliations{
    \textsuperscript{\rm 1}College of Intelligence and Computing, Tianjin University, Tianjin, China\\
    \textsuperscript{\rm 2} Beijing Wenge Technology Co., Ltd, Beijing, China\\
    \{zhang\_jing@tju.edu.cn, pzhang, yxhou\}@tju.edu.cn
}

\begin{document}

\maketitle

\begin{abstract}
Transformer-based large language models exhibit groundbreaking capabilities, but their storage and computational costs are prohibitively high, limiting their application in resource-constrained scenarios. An effective approach is to eliminate redundant model parameters and computational costs while incorporating efficient expert-derived knowledge structures to achieve a balance between compression and performance. Therefore, we propose the \textit{Sememe Entanglement Encoding (SEE)} algorithm. Guided by expert prior knowledge, the model is compressed through the low-rank approximation idea. In Entanglement Embedding, basic semantic units such as sememes are represented as low-dimensional vectors, and then reconstructed into high-dimensional word embeddings through the combination of generalized quantum entanglement. We adapt the Sememe Entanglement Encoding algorithm to transformer-based models of different magnitudes. Experimental results indicate that our approach achieves stable performance while compressing model parameters and computational costs.
\end{abstract}

\section{Introduction}

Transformer-based models have sparked a craze in the deep learning community~\cite{vaswani2017attention,devlin2018bert,achiam2023gpt}. However, the immense scale, computational costs, and storage costs make resource-constrained scenarios hesitant to embrace them. The meaningful research question now is how to reduce the model's parameter count and computational costs without compromising the excellent performance of Transformers~\cite{wu2020lite,ma2019tensorized}.

In the Transformer-based framework, the Embedding layer introduces a significant number of parameters and computational costs~\cite{li2022hypoformer}. For the compression of the component, in the literature, tensor (matrix) decomposition methods based on low-rank approximation have become one of the mainstream approaches due to their ability to eliminate redundant computations~\cite{thakker2020rank}. In the literature, ALBERT~\cite{wu2020lite} reconstructs the model's embedding layer by multiplying two low-rank matrices, reducing the model parameters. Word2Ket~\cite{panahi2019word2ket} adopts the concept of generalized quantum entanglement to combine multiple low-dimensional vectors into a high-dimensional representation. MorphTE~\cite{gan2022morphte} follows the mathematical modeling approach of Word2Ket and further introduces prior knowledge of language structure, decomposing words into morphemes, expressing morphemes with low-dimensional vectors, and reconstructing high-dimensional word representations.

\begin{table}
  \caption{The English word cases of HowNet}
  \begin{center}
  \label{tab:table 1}
  \begin{tabular}{lcc}
    \toprule
     Word &Sense&Sememe   \\
    \midrule
    \multirow{2}{*}{chair}  & chair & ComeTogether, manage, fact\\
    & chair & furniture, sit \\
    \midrule
    \multirow{3}{*}{power}  & power & physical, PhysicsPower \\
    & power & AnimalHuman, Power, politics \\
    & power &math, symbol, Quantity  \\
    & power &country, place, politics  \\
    & power &machine, function, Strength \\
    \bottomrule
  \end{tabular}
  \end{center}
\end{table}

\begin{figure*}[t]
	\centering
	\subfloat[The Word2Ket]{\includegraphics[width=.95\columnwidth]{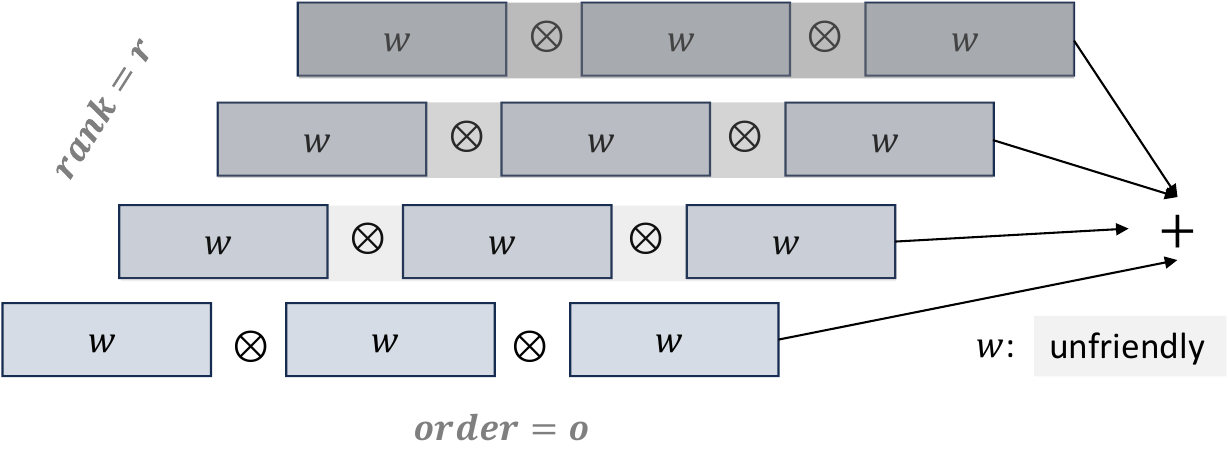}}\hspace{10pt}
	\subfloat[The MorphTE]{\includegraphics[width=.95\columnwidth]{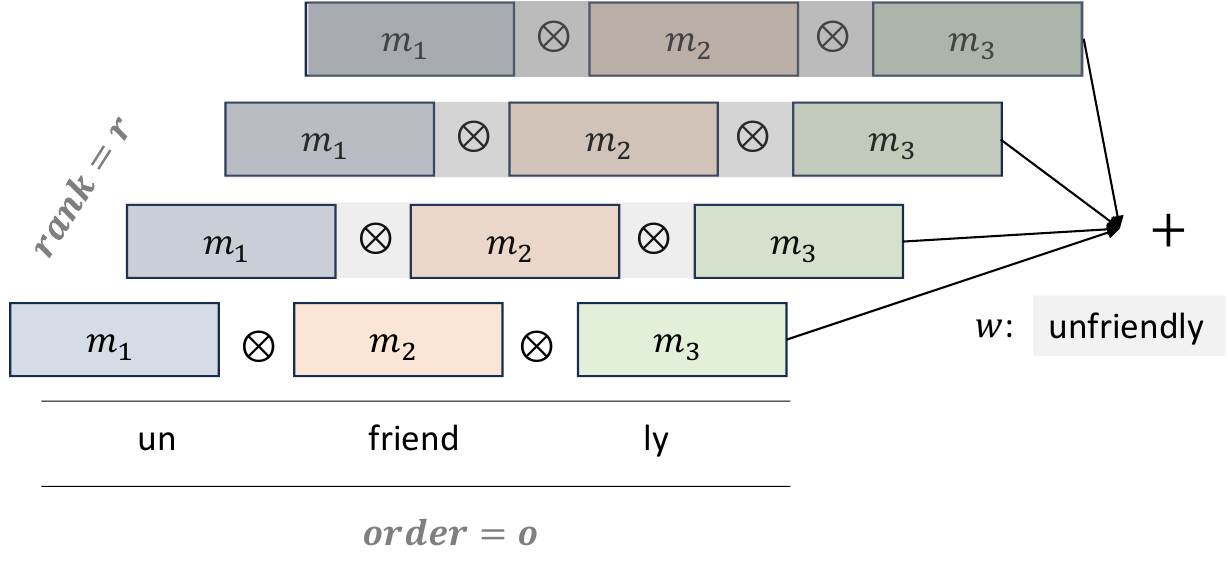}}
	\caption{The diagram of Word2Ket and MorphTE}
    \label{fig.1}
\end{figure*}

However, the above-mentioned methods, when performing low-rank approximations, failed to explicitly assign explicit meaning to the rank, a characteristic that influences the compression rate of the model. Especially in cases where the model has shortcomings in understanding semantic metaphors, blind compression can only harm the performance of the model.

To address the challenges mentioned above, we introduce the semantic cognition system, HowNet~\cite{qi2019openhownet}, abstracted by human experts. This allows the rank to be explicitly specified as the quantity of word semantics.  Specifically, as shown in Table \ref{tab:table 1}, a word like ``power" has five semantics, each identified by different sememes (the smallest indivisible units of meaning). During the initialization of the embedding layer, we adaptively combine this structure with the generalized quantum entanglement modeling method. We represent sememes with low-dimensional vectors, take the tensor product of vectors belonging to the same semantic (i.e., the same rank index), and finally add them to reconstruct the high-dimensional word embedding. This achieves compression while enabling the model to learn semantic interactions at the level of human expert understanding. It is worth noting that relying solely on semantics may not uniquely determine a word. To avoid confusion, we also introduce word structure, namely morpheme groups, as the last rank index. 

From a compression perspective, the number of sememes in HowNet is only around 2000, and the number of morphemes is much smaller than the normal word vocabulary. Each sememe and morpheme are represented by low-dimensional vectors. Compared to the original embedding layer parameters $|V| \times d$, where $|V|$ can reach around 800K, and $d$ usually varies from 512 to 2028, our method achieves compression in both vocabulary size and word dimensionality, resulting in $(|S| + |M|) \times d^{1/o}$, where $|S| + |M|$ represents the sum of morphemes and sememes, and $o$ is the order that identifies the number of sememes for a semantic. To avoid storing information from a basic unit (i.e., sememe and morpheme) in a single low-dimensional vector, we set a basic unit to be identified by m low-dimensional vectors. Thus, the final embedding layer parameter of the proposed method is $(|S| + |M|) \times d^{1/o} \times m$.

In terms of experiments, we conducted evaluations on translation datasets, text matching, and text classification datasets. The results confirmed that the improvement introduced by our method to the Transformer and Phi3-3B large models achieved a balance between performance and the costs of parameters and computations.

\begin{itemize}
    \item We propose the Sememe Entanglement embedding layer, which combines the semantic cognition architecture with the generalized quantum entanglement method. We compress the embedding layer's parameters from both the vocabulary and dimensionality aspects, enhancing the model's ability to learn deep-seated semantic interactions.
    \item Based on the incorporation of Sememe and leveraging the generalized quantum entanglement method as a computational tool, we propose a compression method applicable to both the embedding layer. Experimental results demonstrate that the proposed approach not only achieves parameter and computational cost reduction but also stabilizes performance.
\end{itemize}

\begin{figure*}[t]
	\centering
{\includegraphics[width=2.\columnwidth]{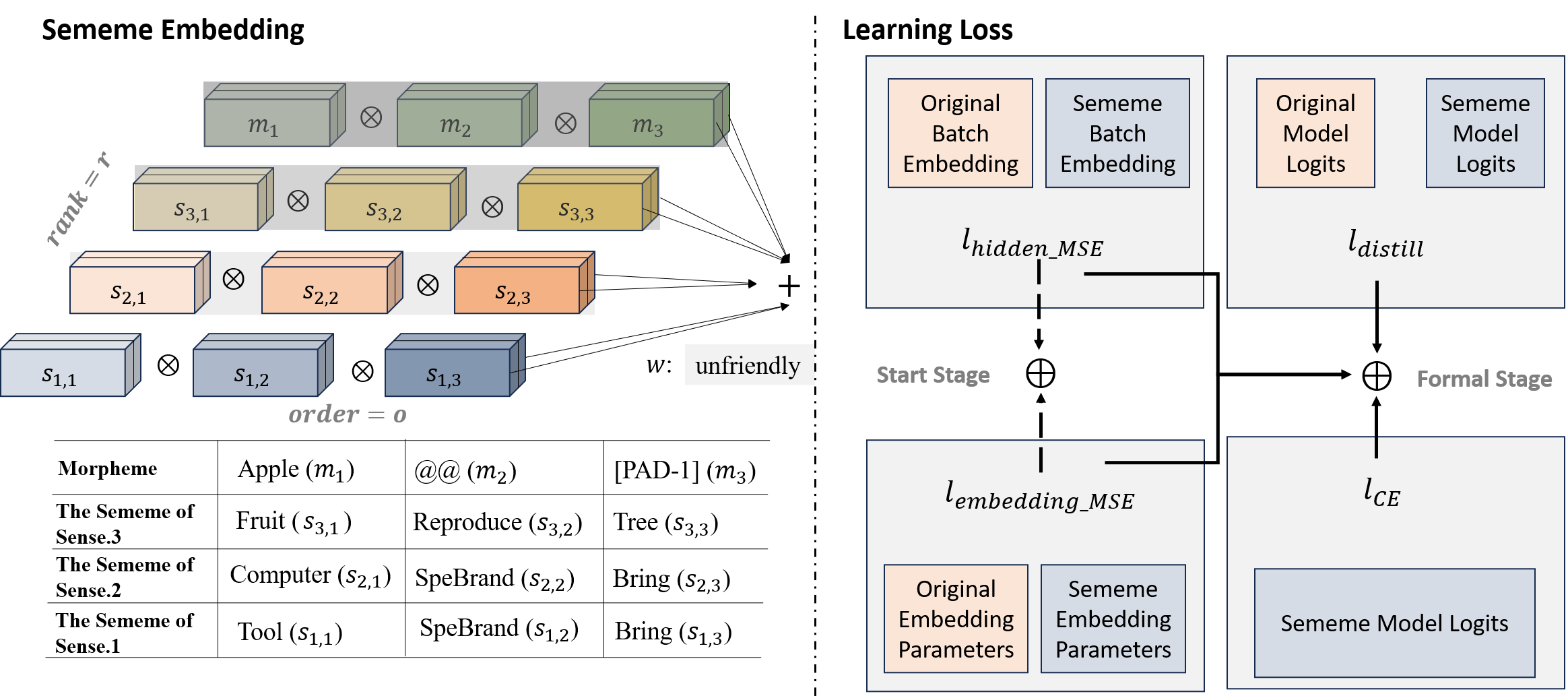}}
	\caption{The Overall Method}
\label{fig.2}
\end{figure*}

\section{Related Work}

Adopting the low-rank approximation method for compression aims to reduce the model parameters while preserving essential information as much as possible. The parameter size of the embedding is $|V|\times d$, where $|V|$ can range from a few thousand to several hundred thousand, and the size of $d$ can also range from 512 to 2028. This results in the embedding layer parameters sometimes occupying a significant proportion, ranging from 20\% to 80\%.

Therefore, based on the concept of low-rank approximation, Word2ket proposes quantum entanglement embedding by integrating the concept of generalized quantum entanglement, to achieve compression~\cite{panahi2019word2ket}. Quantum entanglement refers to the inseparable state of a quantum system composed of particles. From a mathematical perspective, we can define the state of quantum entanglement through the tensor product ($\otimes$) operation~\cite{szalay2015tensor}. Specifically, the quantum state of the composite system cannot be expressed as the direct product state of the quantum state of the subsystem, that is, the composite system has non-separability~\cite{myrvold2011nonseparability}:
\begin{equation}
   |\varphi\rangle \neq |\varphi_{s_1}\rangle \otimes \cdots \otimes |\varphi_{s_N}\rangle
\label{Equ:2.17}
\end{equation}
where $|\varphi\rangle$ is a entanglement system and $|\varphi_{s_1} \rangle$ is a component of its subsystems. Tensor product can be represented as:
\begin{equation}
\mathbf{A}=\mathbf{y} \otimes \mathbf{x} ={\left[ 
\begin{array}{ccc}
x_{1}y_{1} & \cdots & x_{1}y_{n}\\
\vdots & \ddots & \vdots\\
x_{m}y_{1} & \cdots & x_{m}y_{n}
\end{array} 
\right ]},
\label{y1y2}
\end{equation}
where $\mathbf{A}$ is a matrix, which is formed by the tensor product of two vectors $\mathbf{x}$ and $\mathbf{y}$.

As shown in Figure.~\ref{fig.1}(a), word2ket defines $o$ low-dimensional vectors $\mathbf{v}_{jk}$, and the tensor product of these vectors results in a high-dimensional vector (or tensor), referred to as a simple tensor. By ensuring that the rank of the embedding is not 1, meaning the final embedding is formed by adding multiple high-dimensional vectors, word2ket employs a generalized quantum entanglement algorithm to simulate the feature entanglement among low-dimensional vectors.
                         ;['lvfdcgg]
where $\mathbf{e}$ is the embedding of a word. However, on the one hand, this approach can only reduce the dimensionality of the embedding, not the size of the vocabulary. On the other hand, it fails to impart explicit concepts of order and rank.

To address this issue, MorphTE~\cite{gan2022morphte} introduces the concept of morphemes, breaking down words into morphemes as illustrated in Figure~\ref{fig.1}(b) and combining it with Formula 3 to define the order as the number of morphemes in a word. For example,``unfriendly" would be broken down into``un",``friend", and``ly", resulting in an order of 3 for the entangled embedding. This algorithm can learn the commonalities and differences between different words from a linguistic structural perspective, while further reducing the size of the vocabulary. However, both of these algorithms inadequately consider the interaction of learning fine-grained semantic meanings, making it challenging to enhance the model's ability to understand metaphors. Additionally, the meaning of rank is not explicitly defined. Therefore, there are still some shortcomings in balancing model performance and computational costs.

\section{Methodology}
\label{method}

In this section, we propose how to leverage the basic unit of word sense and the basic unit of word morph to construct the quantum entanglement embedding matrix.

\subsection{Embedding Layer Compression}

\subsubsection{The Representation of Basic Units}

Words are not the smallest and most basic units that make up natural language. From the perspective of linguistic morphology, morphemes are the smallest units, and from the perspective of linguistic meaning, primitives are the smallest units. Therefore, morphemes and primitives contain more abstract language understanding information for humans. Based on this, further reconstructing word embedding representations helps large language models perceive the complex metaphorical associations in language more clearly. 

\begin{figure}[t]
	\centering
{\includegraphics[width=.98\columnwidth]{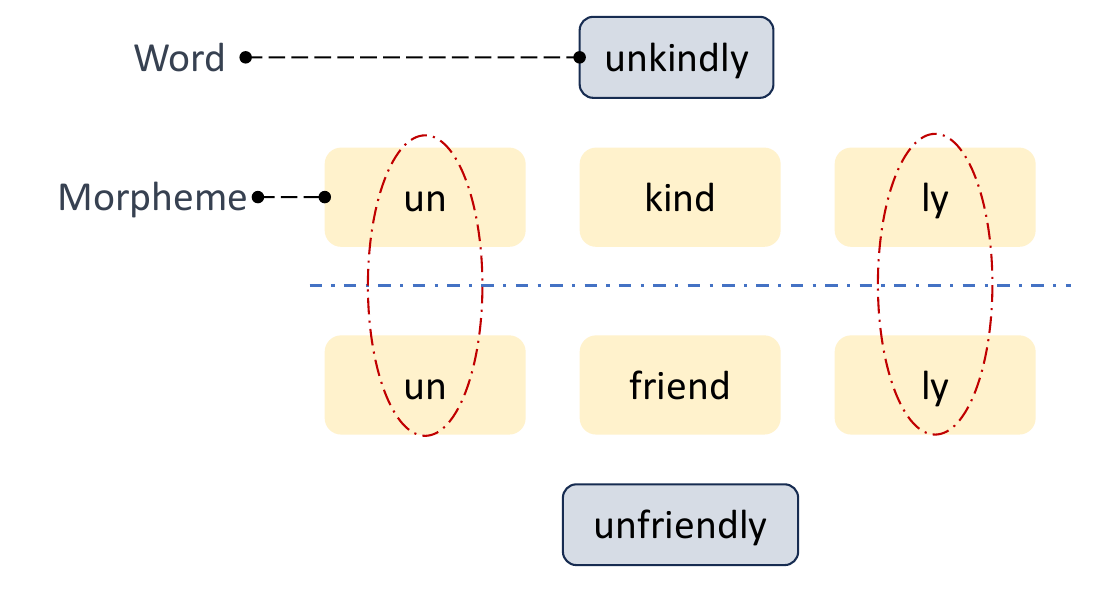}}
	\caption{Example of Morpheme Decomposition}
\label{fig.3}
\end{figure}

\textbf{Morpheme}  As shown in Figure.~\ref{fig.3}, a token ``unfriendly'' can be split as multiple morpheme ``\{un, friend, ly\}'' and a token ``unkindly'' is split as ``\{un, kind, ly\}''. Between the two tokens, there are common morphemes ``un'' and ``ly'', so using morphemes as the basic units to construct token embeddings can model the underlying abstract structural relationships between tokens.

It is worth noting that the addition of morphemes serves to uniquely determine a word structurally.

\textbf{Sememe} In different contexts, a token can have varied meanings, and considering human creativity, abstract meanings in language are often utilized to create various metaphors. Therefore, linguistic experts, drawing on their cognitive expertise, have summarized the abstract meanings of hundreds of thousands of vocabulary items in both Chinese and English to form the sememe library, that is HowNet. In this context, ``sememe'' refers to the smallest semantic units that is the most basic, indivisible minimal semantic unit\cite{qi2019openhownet}, facilitating a deeper and more thorough understanding of linguistic semantics.

\begin{figure}[t]
	\centering
{\includegraphics[width=.98\columnwidth]{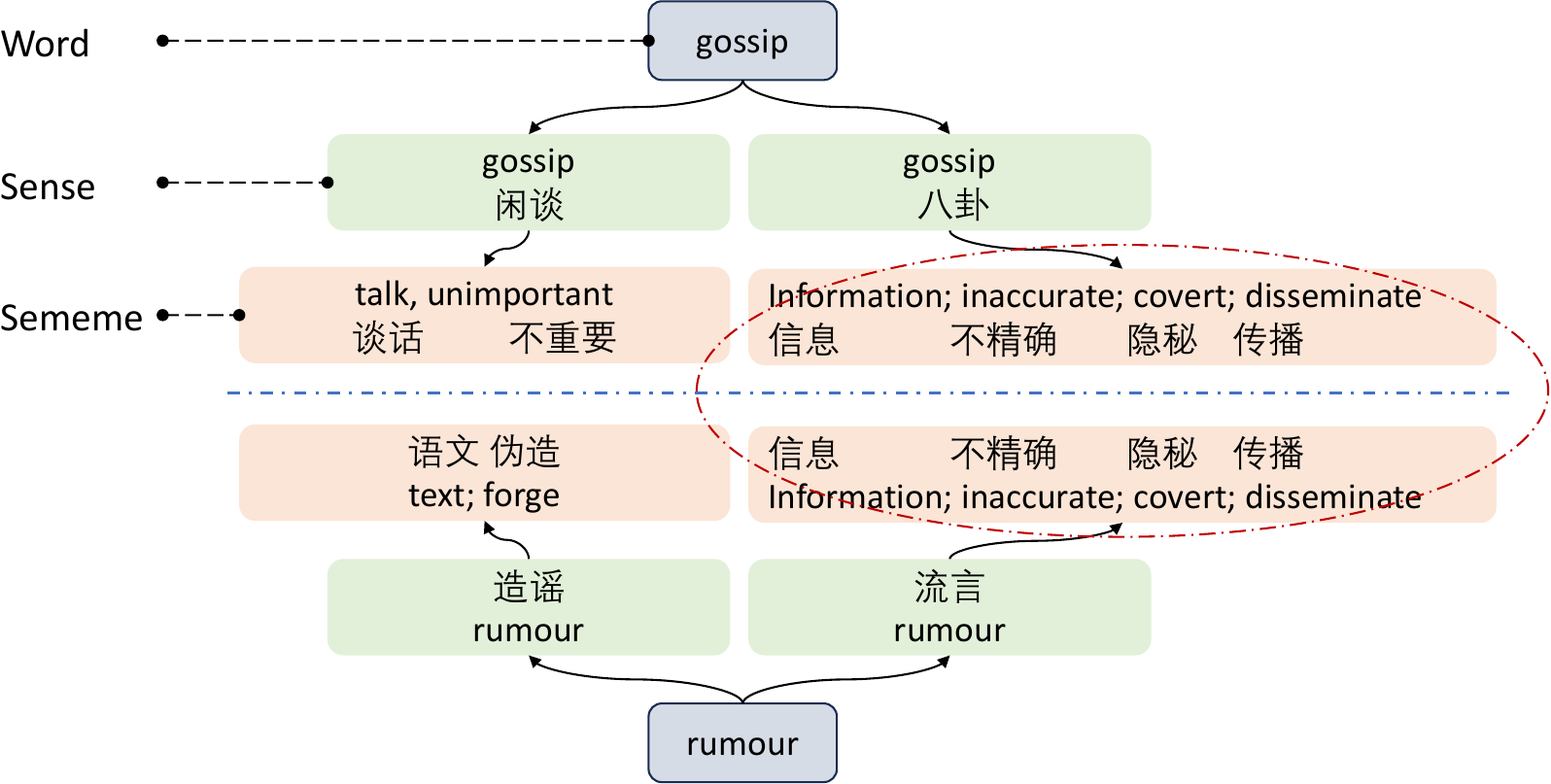}}
	\caption{Example of Sememe Decomposition}
\label{fig.4}
\end{figure}

As shown in Figure.~\ref{fig.4}, there are two words ``gossip'' and ``rumour'', it can be observed that, whether from a semantic or visual perspective, these two words do not have an apparent correlation. However, through sememe, a connection between these two words can be established at a more detailed level. Therefore, employing semantic radicals to construct quantum entanglement embeddings can further enhance the model's ability to comprehend semantics at a deeper level.

Based on the above analysis, we define a word as having $r\times o$ basic units, specifically $1\times o$ morphemes, $r-1$ senses, with each sense containing $o$ sememe.

\subsubsection{Quantum Entanglement Embedding}

Broadly speaking, quantum entanglement is a system with complex interactions that cannot be directly represented by the tensor product of internal components, indicating unconstrained entangled relationships among components. Therefore, this paper naturally regards the embedding representation of words as a complex system, with components being basic semantic units and basic morphological units. The paper utilizes generalized quantum entanglement to model the interactive relationships among basic units.
\begin{equation}
    \mathbf{e} = \sum_{i}^{m}\sum_{j}^{r}\otimes_{k}^{o}\mathbf{v}_{jk,i}
\end{equation}
where $\mathbf{v}_{ij, m}$ is a vector with $d^{1/o}$, and $d$ is the embedding dimension of original model. As shown in Figure~\ref{fig.2}, the $o$ is the order, $r$ is the rank, 
each rank represents a semantic aspect or structural aspect of the word, each of which includes $o$ morpheme or $o$ sememe, respectively.  

In addition, to avoid the issue of insufficient information capacity caused by excessively low embedding dimensions for basic units, we set each basic unit to be represented by $m$ sets of low-dimensional vectors. After entanglement computations transform these low-dimensional vectors into high-dimensional vectors, the $m$ sets of high-dimensional vectors are then summed.

Thus, the entire workflow of embedding layer can be formalized as follows:
\begin{equation}
    \mathrm{Embed}(w_e) = \sum_{i}^{m}\sum_{j}^{r}f_{i}(I_{e,j1})\otimes \cdots \otimes f_{i}(I_{e,jo})
    \label{eq.3.1.2.2}
\end{equation}
where the function $f_i(:)$ is designed to transform the indices of the basic semantic units (sememe and morpheme) of a word into corresponding low-dimensional vectors. The $m$ represents each index with $m$ low-dimensional vectors to store more information.

It's worth noting that our method does not consume extra computational time during the inference process since we still look up embeddings through word indices, and the embedding calculations, as shown in Equation 3.1.2.2, are completed during the model initialization.

\subsubsection{Difference between SEE and Other Methods}

First, compared to the Word2Ket model, we implemented dual compression for both vocabulary size and vocabulary dimension, similar to MorphTE. In contrast to MorphTE, our algorithm achieves rank flexibility. Rank is no longer a black-box concept; it is endowed with precise semantic meaning. With this foundation, setting rank to 1 or 100 has no impact on the parameter count of the embedding.

Additionally, since sememe represent the smallest units of semantic, HowNet has only slightly over 2000 sememe. It has annotated concepts represented by several hundred thousand Chinese and English words with these sememe~\cite{qi2019openhownet}. Therefore, compared to MorphTE, our vocabulary size has increased by only two thousand. With the achieved rank flexibility and We typically set the value of $m$ to be small, we can achieve higher degrees of compression.

Finally, we introduce abstract knowledge structures extracted by experts to construct embeddings enables a deeper semantic interaction for the model. Compression at the embedding layer is realized through guided human prior knowledge.

\begin{table*}[t]
  \caption{Comparison experimental results based on different embedding compression methods. The compression ratio is reported in the format B(C×), and BLEU is used as the evaluation metric for translation tasks. - indicates that the method cannot achieve approximately the same compression ratio as other methods.}
  \begin{center}
  \label{tab:table 2}
  \begin{tabular}{l|cccc|cccc}
    \toprule
    &\multicolumn{4}{|c}{WMT17 ZH-EN} &\multicolumn{4}{|c}{IWSLT17 ZH-EN}    \\

    \midrule
    &\multicolumn{8}{|c}{BLEU4} \\
        \midrule
        Transformer$+$ & \multicolumn{4}{c}{15.16}  & \multicolumn{4}{|c}{21.90}  \\
    \midrule
     &10$\times$ &20$\times$ &40$\times$ &80$\times$ &10$\times$ &20$\times$ &40$\times$ &80$\times$\\
    \midrule
    Matrix  & 13.39  &11.48  &6.50  &0.35  & 19.56 & 17.28  &12.31 &1.47   \\
    TT  & 12.13  &8.54  &--  &--  & 18.43 & 14.03  &--  &--  \\
    Word2ket  & 14.28  &--  &--  &--  & 20.86  &  -- &--  &--   \\
    MorphTE  &15.00   & 14.56 &  13.43 & 12.2  & 20.91 & 20.63 & 19.90 &18.70 \\
    \midrule
    Ours (S-Embed)  & 14.74 & 14.09 &13.32 &\textbf{12.62} & \textbf{21.50} & \textbf{20.76} &\textbf{19.91} & \textbf{19.36}\\
    \bottomrule
  \end{tabular}
  \end{center}
\end{table*}

\begin{table*}[t]
    \centering
    \caption{Statistics of translation tasks. $|V|$ is the size of the word vocabulary. $S$ and $E$ are the model parameters and the embedding parameters, respectively. P is the proportion of the embedding layer to the total parameters. $|M|$ is the size of the sememe vocabulary}
    \begin{tabular}{ccc|ccc|cc|cc|cc|cc}
        \toprule
         \multicolumn{3}{c|}{Original Model} &\multicolumn{11}{|c}{Sememe Model}\\
         \midrule
         $|V|$   &$S / E$ & P  &$|M|$ & $o / r$ & $d^{1/o}$ & ratio & $m$ & ratio &  $m$ & ratio &$m$  & ratio  &$m$\\
         \midrule
         46272 &  55.2M / 23.7M  &42.9\% & 16325  & 3 / 5 & 8  &10$\times$ &18 &20$\times$ &9  &40$\times$ &4 &80$\times$ &2 \\
        \bottomrule
    \end{tabular}
    \label{tab:table 3}
\end{table*}

\subsection{The Finetuning Method of Embedding Layer}

To use SEE on a general large model, this paper proposes a multi-stage, multi-granularity distillation method. In the initial stage, we first apply the MSE loss on the embedding parameters and embedding hidden states to bring the source large model and the student small model with SEE closer together. In the subsequent formal training stage, we incorporate logits-based distillation loss and the Cross-Entropy loss from the sememe encoding model.

The mean squared error (MSE) loss between the embedding layer parameters of the original model and sememe embedding model:

\begin{equation}
\mathcal{L}_{\text{embedding\_MSE}} = \frac{1}{n} \sum_{i=1}^{n} (\mathbf{e}_{\text{ori}}^{(i)} - \mathbf{e}_{\text{sem}}^{(i)})^2
\end{equation}
where \(\mathbf{e}_{\text{ori}}^{(i)}\) and \(\mathbf{e}_{\text{sem}}^{(i)}\) are the embedding parameters of the original and the compression model at the \(i\)th position, and \(n\) is the number of embedding parameters. The MSE loss between the embedding hidden states of the original and small models:

\begin{equation}
\mathcal{L}_{\text{hidden\_MSE}} = \frac{1}{m} \sum_{j=1}^{m} (\mathbf{h}_{\text{ori}}^{(j)} - \mathbf{h}_{\text{sem}}^{(j)})^2
\end{equation}
where \(\mathbf{h}_{\text{ori}}^{(j)}\) and \(\mathbf{h}_{\text{sem}}^{(j)}\) represent the hidden states of the teacher and student models at the \(j\)th time step, and \(m\) is the number of time steps. The distillation loss between the original model and the compression model, is used to make the  output of small model resemble the the output of original LLM as closely as possible:
\begin{equation}
\mathcal{L}_{\text{distill}} = \text{KL}\left(\text{Softmax}\left(\frac{\mathbf{z}_{\text{ori}}}{T}\right) \,\Bigg\|\,
\text{Softmax}\left(\frac{\mathbf{z}_{\text{sem}}}{T}\right)\right)
\end{equation}
where \(\mathbf{z}_{\text{ori}}\) and \(\mathbf{z}_{\text{sem}}\) are the logits of the teacher and student models, respectively, and \(T\) is the temperature parameter to soften the logits. The cross-entropy loss for the compression sememe model on the tasks:
\begin{equation}
\mathcal{L}_{\text{CE}} = -\frac{1}{N} \sum_{i=1}^{N} \sum_{c=1}^{C} y_{i,c} \log(\hat{y}_{i,c})
\end{equation}
where \(N\) is the number of samples, \(C\) is the number of classes, \(y_{i,c}\) is the one-hot encoded true label for the \(i\)th sample and \(c\)th class, and \(\hat{y}_{i,c}\) is the predicted probability of the SEE+LLM model for the \(i\)th sample and \(c\)th class.

Finally, the total loss can be formulated as a weighted sum of these individual losses:

\begin{equation}
\mathcal{L} = \alpha \mathcal{L}_{\text{distill}} + \beta \mathcal{L}_{\text{embedding\_MSE}} + \gamma \mathcal{L}_{\text{hidden\_MSE}} + \mathcal{L}_{\text{CE}}
\end{equation}
where $\alpha$, $\beta$ and $\gamma$  are the weighting coefficients for each loss component.

\section{Experiments}

In the experiment, we need to verify the following two questions.
\begin{itemize}
    \item In the Transformer-based architecture, the proposed embedding compression methods aim to maintain good performance for the model under high compression ratios and low computational costs.
    \item The constructed Sememe Entanglement Embedding method is assessed for its superiority over other algorithms incorporating sememe information. 
    \item For large language models, the proposed methods are evaluated for their ability to maintain robust performance at high compression rates and low computational costs.
\end{itemize}

\subsection{Baselines}

To evaluate the first issue, we chose \textbf{Word2Ket}~\cite{panahi2019word2ket} and \textbf{MorphTE}~\cite{gan2022morphte} embedding compression algorithms as baselines. Word2Ket is a quantum entanglement embedding for compressing the dimension size of word embedding. 
MorphTE achieves dual compression of vocabulary and dimensions on the basis of word2ket by introducing morphemes. In addition, we chose Tensor-Train (\textbf{TT}) decomposition~\cite{oseledets2011tensor} and low–rank matrix factorization (\textbf{Matrix})~\cite{mnih2007probabilistic}. The compression rates of these two methods are closely related to the rank.

To evaluate the second issue,  \textbf{MorphTE} and \textbf{MorphLSTM}~\cite{gan2022morphte} and \textbf{SememeLSTM} incorporate morphemes into embeddings. We validate the effectiveness of our model's embeddings for semantic and metaphor understanding on these three baselines. 

To evaluate the third issue, we chose \textbf{Phi-3B}~\cite{abdin2024phi} large language model as our baseline. We improved the proposed algorithm to obtain Phi+SEE, and conducted comparative experiments with the source Phi3 model.

\subsection{Tasks, Datasets, and Metrics}
For evaluating the first question, we conduct experiments on the  \textbf{WMT17 ZH-EN}~\cite{sennrich2017university} and \textbf{IWLST2017 ZH-EN}~\cite{cettolo2017overview} translation task. We use WMT17 News Commentary translation dataset which consists of about 320K sentence pairs The data is processed by the BPE~\cite{sennrich2015neural}. IWLST2017 ZH-EN has 232K sentence pairs in train set, 888 sentence pairs in valid set and 8.58K sentence pairs in test set. The shared vocabulary size for source and target is 40K, while the source embedding and target embedding are not shared. The performance is measured by case-sensitive tokenized BLEU~\cite{papineni2002bleu} for all translation
tasks. 

We used our proposed structure on two popular zero-shot generation tasks, including ARC-Challenge~\citep{clark2018think}, ARC-Easy~\citep{clark2018think} with higher accuracy, indicating that Mooe has a stronger parameter fine-tuning ability to handle downstream tasks.

\subsection{Experimental Setting}
For machine translation tasks, we chose Transformer with the implementation of Fairseq. For all WMT17 ZH-EN and IWSLT ZH-EN datasets, the Transformer consists of 6-layer encoder and 6-layer
decoder with 512 embedding size, 1024 feed-forward network (FFN) size. It is trained with a batch
size of 4096 tokens on an NVIDIA Tesla V100 GPU. The rank is set to 5 and the order is set to 3, which means that a word contains a set of morphemes and four sets of sememes, with each set having 3 units.

We used Phi3 as our testing model, Phi3 has 3B parameters and 32 layers, the dimension of embedding is 3072, we use them to test the best model settings on models of different sizes. We use Adam as the optimizer with a learning rate of $4 \times 10^{-5}$ for fine-tuning downstream tasks and set the batch size to 32.

\subsection{Main Results}

To validate the compression effectiveness of Sememe Entanglement Embedding (SEE), experiments were conducted on English-Chinese translation using Transformer models. As shown in Table \ref{tab:table 2}, we performed high compression ratios of 10x, 20x, 40x, and 80x on the embedding layer. Notably, SEE achieves good results even with 80x compression. For example, on the IWSLT dataset, SEE only decreased by 2.54 BLEU compared to the original model, while MorphTE decreased by 3.2 BLEU. MF compression methods struggled to maintain the model's basic performance, and TT decomposition and Word2KET decomposition had difficulty achieving high compression ratios.

For lower compression ratios, such as 10x and 20x, the proposed method generally maintains stable model performance. On the IWSLT dataset, with 10x compression, the model's performance only decreased by 0.4 BLEU points. This suggests that the proposed method combines the smallest semantic units with the smallest structural units of tokens through a generalized quantum entanglement approach, providing the model with more explicit expert knowledge. This not only eliminates redundant parameters but also allows the incorporation of effective external information, resulting in strong performance in compression scenarios.

Overall, as shown in Table \ref{tab:table 3}, which details the configuration for multi-task translation, when the proportion of parameters (P) in the embedding part is around 42.9\%, the Sememe approach can still maintain good model performance even under more than 10-fold compression.

\subsection{Compression on Phi3}

To validate the effectiveness of the proposed compression and distillation methods in large language models, we conducted experiments on the Phi3-8B model as well as the ARC-c and ARC-e datasets. Specifically, with rank set to 5, order set to 3 ($d^{1/o}=15$), and m set to 10, the introduction of SEE compressed the embedding layer parameters of the Phi3 model by nearly five times.

Regarding training, we performed the distillation in the start stage during the first two epochs, followed by formal stage training. Additionally, due to the special nature of embedding compression, unlike FFN and Attention models, only token embeddings that appeared in the fine-tuning training set can be learned during fine-tuning. This paper preliminarily demonstrates the feasibility of embedding SEE in large models, and thus selected some data from the training set as a test set.

The test results show that with a 5x compression, the model's performance on the ARC-c dataset decreased by only 0.9\%, and on the ARC-e dataset by 0.2\%, with an average reduction of 0.6\% in multi-task learning performance. This experimental result indicates the feasibility of introducing sememe compression and multi-granularity distillation in large models.

\begin{table}
  \caption{Comparison experiments on the Phi3-3B model}
  \begin{center}
  \label{tab:table 4}
  \begin{tabular}{l|c|ccc}
    \toprule
\textbf{model}&\textbf{parameter}&\textbf{ARC-c}&\textbf{ARC-e}&\textbf{avg.}\\
    \midrule
    \midrule
    \multirow{2}{*}{Phi3} &100\% &1.0 &1.0 &1.0\\
 &22.7\%&0.991 &0.998 &0.994\\
    \bottomrule
  \end{tabular}
  \end{center}
\end{table}

\subsection{Effectiveness of Introducing Sememes}

To verify the effectiveness of introducing Sememe in a non-compressed state, experiments were conducted on the WMT17 ZH-EN and IWSLT17 ZH-EN translation datasets. Specifically, instead of constructing a vocabulary through entangled vectors, the Sememe and morphemes were introduced into the model by inputting the high-dimensional vectors of the morphemes or Sememe related to the current token into the LSTM model to encode the high-dimensional representation of the token.

As shown in Table \ref{tab:table 5}, compared to the source model, SememeLSTM achieved a 1.9\% improvement on the WMT task and a 1.1\% improvement on the IWSLT dataset. This indicates that Sememe enhances the model's performance by introducing finer-grained semantic and language structure information. Compared to the MorphLSTM method, the proposed model showed a 5.5\% improvement on the WMT dataset and a 4.8\% improvement on the IWSLT dataset. This suggests that simply adding structural information of the language might disrupt the overall semantic meaning of the token. In contrast, the Sememe method, through its unique combinatory structure, introduces more specialized knowledge, thereby enhancing the model's expressive capability."

\begin{table}
  \caption{Validation of the effectiveness of sememes}
  \begin{center}
  \label{tab:table 5}
  \begin{tabular}{l|c|c}
    \toprule
    &\multicolumn{1}{|c}{WMT17 ZH-EN} &\multicolumn{1}{|c}{IWSLT17 ZH-EN}    \\

    \midrule
    &\multicolumn{2}{|c}{BLEU4} \\
        \midrule
        Transformer  &15.16 &21.90  \\
        \midrule
        MorphLSTM$+$ & 14.65  & 21.14 \\
        SememeLSTM$+$ & 15.45  &  22.15 \\
    \bottomrule
  \end{tabular}
  \end{center}
\end{table}

\subsection{Performance analysis}

As shown in Figure 5, sensitivity tests were conducted on rank and order. The red squares represent the number of parameters, the blue line represents the BLEU score on WMT, and the black line represents the BLEU score on IWSLT. The specific settings are as follows: the default setting for rank is 5, order is set to 3, and m is 9. In the first figure, the rank size was adjusted, and the experiment shows that the parameters of the embedding module remain constant, with the embedding compressed by 20 times compared to the source model. This indicates that the rank has a minimal impact on model parameters. Moreover, the BLEU results suggest that in tasks that are particularly prone to overfitting, the rank setting significantly affects model performance. Therefore, when experimenting with specific models, the rank setting needs to be adjusted based on the training conditions.

In the experiments with order, when m and rank remain unchanged and only the order varies, the parameters of the embedding layer gradually decrease. Comparing the trend in BLEU scores, we also found that the order setting significantly affects model performance in translation tasks. This is because when the order is too small, the granularity of token structure and semantic decomposition becomes too coarse. As the order increases, the dimension of each sememe unit decreases significantly, which may lead to insufficient information storage. Therefore, the experimental results show a high sensitivity to the order setting.

\begin{figure}[t]
    \begin{subfigure}[Rank vs BLEU and Embedding Parameters]{\textwidth}
\includegraphics[width=0.45\textwidth]{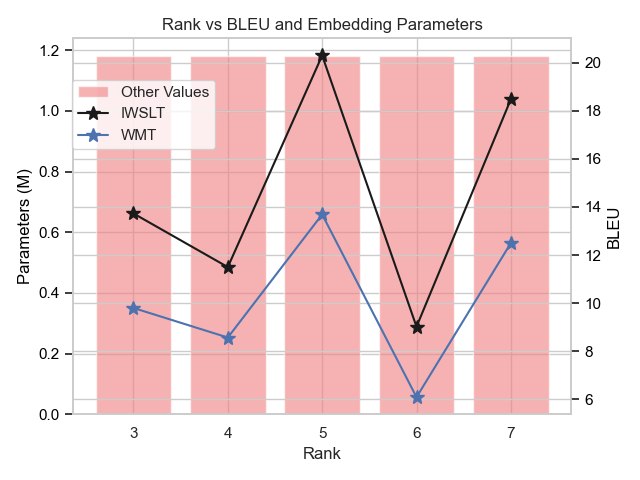} 
        \label{fig:sub1}
    \end{subfigure}
    \begin{subfigure}[Order vs BLEU and Embedding Parameters]{\textwidth}
\includegraphics[width=0.45\textwidth]{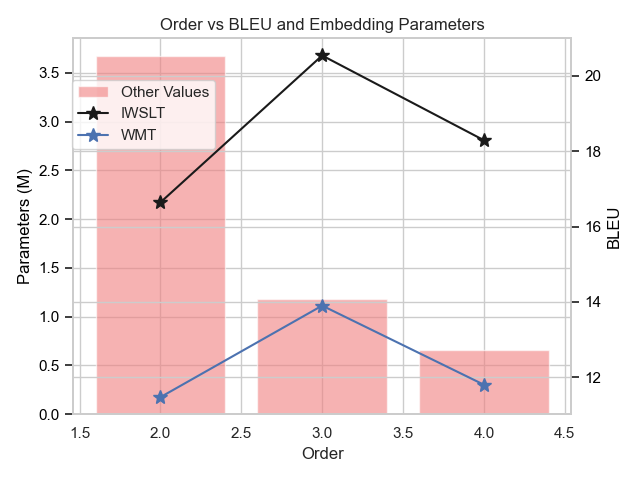} 
        \label{fig:sub2}
    \end{subfigure}
    \caption{Hyperparameter Analysis}
    \label{fig:main}
\end{figure}

\section{Conclusion}

We propose the Sememe Entanglement Encoding method. Sememe is the smallest semantic units, using low-dimensional vectors, and we model the interaction between sememe representations based on generalized quantum entanglement calculations, reconstructing them into high-dimensional word embedding vectors. For pre-trained large models, this paper also proposes a distinctive distillation mechanism. At the initial stage, MSE loss is applied to the embedding hidden states and embedding parameters between the teacher and student modules. Subsequently, the model's performance is further enhanced by incorporating the student's model loss and KL distillation loss. In terms of experiments, we validate our method on the WMT17 ZH-EN and IWSLT17 ZH-ZH translation datasets based on the Transformer architecture. Our approach achieves substantial compression while maintaining performance. The effectiveness of the proposed algorithm is further validated on the 1B model.

\bibliography{aaai25}
\end{document}